\documentclass[10pt,twocolumn,letterpaper]{article}
\usepackage{cvpr}              
\usepackage{graphicx}
\usepackage{amsmath}
\usepackage{amssymb}
\usepackage{booktabs}
\usepackage{multirow}
\usepackage[pagebackref,breaklinks,colorlinks]{hyperref}
\usepackage{bbm}
\usepackage[capitalize]{cleveref}

\newcommand{\beginsupp}{
        \setcounter{table}{0}
        \renewcommand{\thetable}{S\arabic{table}}
        \setcounter{figure}{0}
        \renewcommand{\thefigure}{S\arabic{figure}}
        \setcounter{section}{0}
        \renewcommand{\thesection}{\Alph{section}}
     }

\crefname{section}{Sec.}{Secs.}
\Crefname{section}{Section}{Sections}
\Crefname{table}{Table}{Tables}
\crefname{table}{Tab.}{Tabs.}
\def\cvprPaperID{2828} 
\def\confName{CVPR}
\def\confYear{2023}
\usepackage[svgnames,table]{xcolor} %
\definecolor{rgb1}{RGB}{214,  38, 40}   
\definecolor{rgb2}{RGB}{43, 160, 4}     
\definecolor{rgb3}{RGB}{158, 216, 229}  
\definecolor{rgb4}{RGB}{114, 158, 206}  
\definecolor{rgb5}{RGB}{204, 204, 91}   
\definecolor{rgb6}{RGB}{255, 186, 119}  
\definecolor{rgb7}{RGB}{147, 102, 188}  
\definecolor{rgb8}{RGB}{30, 119, 181}   
\definecolor{rgb9}{RGB}{160, 188, 33}   
\definecolor{rgb10}{RGB}{255, 127, 12}  
\definecolor{rgb11}{RGB}{196, 175, 214} 
\usepackage{pifont}
\newcommand{\cmark}{\ding{51}}%

\ifdefined \GramaCheck
  \newcommand{\CheckRmv}[1]{}
  \renewcommand{\eqref}[1]{Eq. (1)}
  \renewcommand{\equref}[1]{Eq. (1)}
  \renewcommand{\figref}[1]{Figure 1}
  \renewcommand{\tabref}[1]{Table 1}
\else
  \newcommand{\CheckRmv}[1]{#1}
  \renewcommand{\eqref}[1]{Eq.~(\ref{#1})}
\fi
\graphicspath{{./figures/}}
\DeclareGraphicsExtensions{.pdf,.jpg,.png,.bmp}

\begin{document}
\title{Semantic Scene Completion with Cleaner Self}
\author{Fengyun Wang$^1$ \ Dong Zhang$^2$ \ Hanwang Zhang$^3$ \ Jinhui Tang$^1$\thanks{Corresponding author.} ~\ Qianru Sun$^4$\\
{\small $^1$School of Computer Science and Engineering, Nanjing University of Science \& Technology}\\
{\small $^2$The Hong Kong University of Science \& Technology \ $^3$Nanyang Technological University \ $^4$Singapore Management University}\\ 
{\small E-mail: \{fereenwong, jinhuitang\}@njust.edu.cn; dongz@ust.hk; hanwangzhang@ntu.edu.sg; qianrusun@smu.edu.sg}}
\maketitle
\begin{abstract}
Semantic Scene Completion (SSC) transforms an image of single-view depth and/or RGB 2D pixels into 3D voxels, each of whose semantic labels are predicted. SSC is a well-known ill-posed problem as the prediction model has to ``imagine'' what is behind the visible surface, which is usually represented by Truncated Signed Distance Function (TSDF). Due to the sensory imperfection of the depth camera, most existing methods based on the noisy TSDF estimated from depth values suffer from  1) incomplete volumetric predictions and 2) confused semantic labels. To this end, we use the ground-truth 3D voxels to generate a perfect visible surface, called TSDF-CAD, and then train a ``cleaner'' SSC model. As the model is noise-free, it is expected to focus more on the ``imagination'' of unseen voxels. Then, we propose to distill the intermediate ``cleaner'' knowledge into another model with noisy TSDF input. In particular, we use the 3D occupancy feature and the semantic relations of the ``cleaner self'' to supervise the counterparts of the ``noisy self'' to respectively address the above two incorrect predictions. Experimental results validate that our method improves the noisy counterparts with $3.1\%$ IoU and $2.2\%$ mIoU for measuring scene completion and SSC, and also achieves new state-of-the-art accuracy on the popular NYU dataset. The code is available at this \href{https://github.com/fereenwong/CleanerS}{link}.
\end{abstract}
\section{Introduction}
\label{sec:intro}
3D scene understanding is an important visual task for many practical applications, \eg, robotic navigation~\cite{garg2020semantics} and augmented reality~\cite{tanzi2021real}, where the scene geometry and semantics are two key factors to the agent interaction with the real world~\cite{hou2020revealnet,zhang2021self}.
However, visual sensors can only perceive a partial world given their limited field of view with sensory noises~\cite{roldao20223d}. Therefore, an agent is expected to leverage prior knowledge to estimate the complete geometry and semantics from the imperfect perception.
Semantic Scene Completion (SSC) is designed for such an ability to infer complete volumetric occupancy and semantic labels for a scene from a single depth and/or RGB image~\cite{song2017semantic,roldao20223d}.

Based on an input 2D image, the 2D$\rightarrow$3D projection is a vital bond for mapping 2D perception to the corresponding 3D spatial positions, which is determined by the depth value~\cite{chen2019learning}.
After this, the model recovers the visible surface in 3D space, which sheds light on completing and labeling the occluded regions~\cite{song2017semantic,li2019rgbd}, because the geometry of the visible and occluded areas is tightly intertwined.
For example, you can easily infer the shapes and the semantic labels when you see a part of a ``{chair}'' or ``{bed}''.
Thus, a high-quality visible surface is crucial for the SSC task.
\begin{figure*}[!t]
\centering
\includegraphics[width=.96\textwidth]{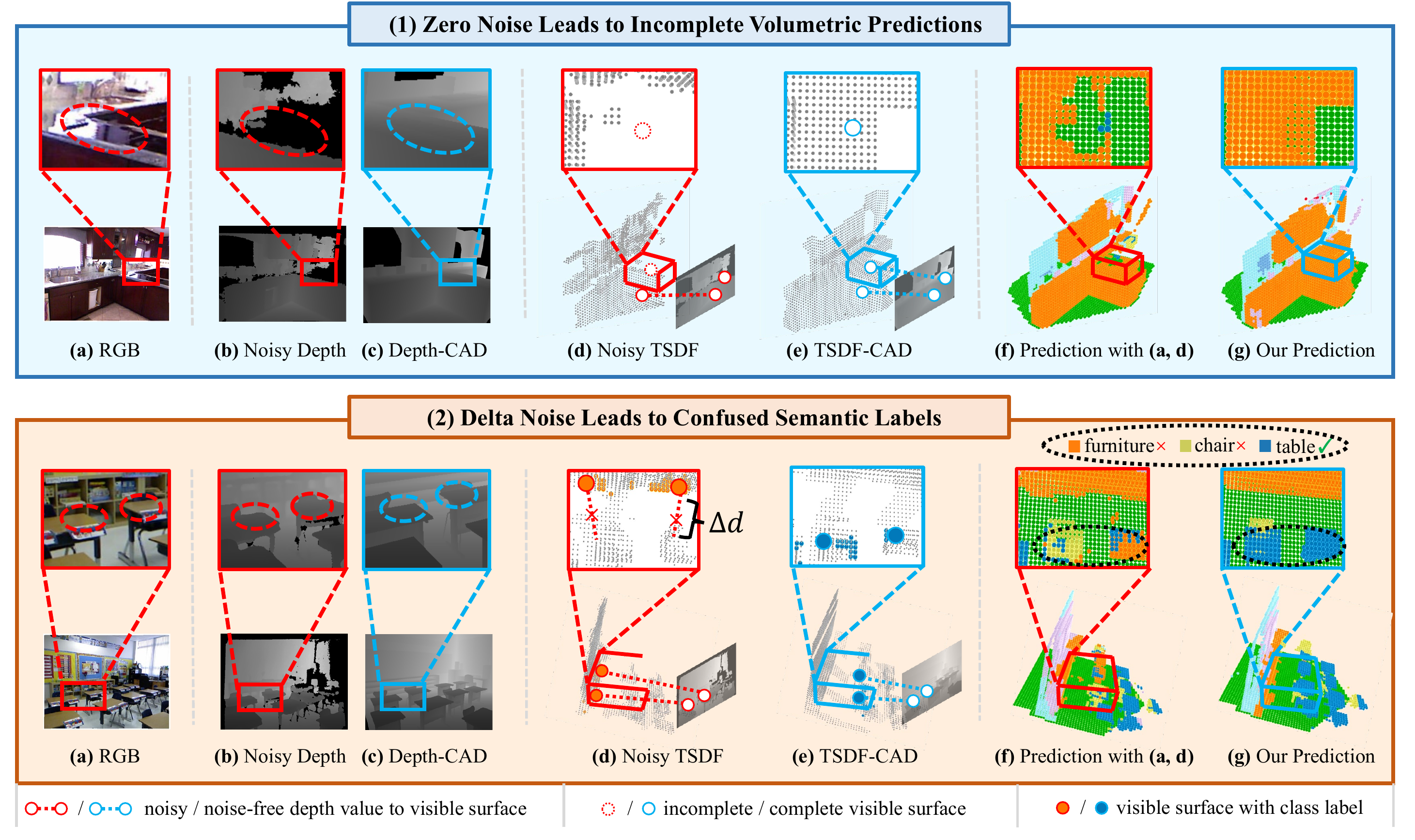}
\vspace{-4mm}
\caption{The existing depth noises can be roughly categorized into: 1) \textbf{zero noise} and 2) \textbf{delta noise}. By \textbf{zero noise}, we mean that when the depth camera cannot confirm the depth value of some local regions, it fills these regions with zeroes, leading to the problem of incomplete volumetric predictions. 
By \textbf{delta noise}, we mean the inevitable deviation (\ie, $\Delta d$) of the obtained depth value due to the inherent quality defects of the depth camera, which leads to the problem of confused semantic labels in the final 3D voxels.
In the above blocks, the pairwise subfigures (e.g., (d) and (e))
show the cases of ``with noise'' and ``without noise'' on the left and right, respectively.}
\label{fig:problem}
\end{figure*}

However, due to the inherent imperfection of the depth camera, the depth information is quite noisy, what follows is an imperfect visible surface that is usually represented by  Truncated Signed Distance Function (TSDF)~\cite{song2017semantic}. In general, the existing depth noises can be roughly categorized into the following two basic types:

\noindent\textbf{1) Zero Noise}. This type of noise happens when a depth sensor cannot confirm the depth value of some local regions, it will fill these regions with zeroes~\cite{fu2012kinect,nguyen2012modeling}. Zero noise generally occurs on object surfaces with reflection or unevenness~\cite{mallick2014characterizations}. 
Based on zero noise, the visible surface will be incomplete after the 2D-3D projection via TSDF~\cite{roldao20223d}, so the incomplete volumetric prediction problem may occur in the final 3D voxels. 
For example, as shown in the upper-half of Figure~\ref{fig:problem}, for the input RGB ``{kitchen}'' image, the depth value of some parts of the ``{cupboard}'' surface (marked with the red dotted frames) (in (b)) is set to zero due to reflections. Based on this, both the visible surface (in (d)) and the predicted 3D voxels (in (f)) appear incomplete in reflective regions of this ``{cupboard}''. Our method uses the perfect visible surface (in (e)) generated by the noise-free ground-truth depth value (in (c)) as intermediate supervision in training , which helps the model to estimate ``{cupboard}'' 3D voxels in inference even with the noisy depth value as input.

\noindent\textbf{2) Delta Noise}. This type of noise refers to the inevitable deviation of the obtained depth value due to the inherent quality defects of the depth camera~\cite{mallick2014characterizations}, \ie, the obtained depth value does not match the true depth value. Delta noise shifts the 3D position of the visible surface, resulting in the wrong semantic labels, such that the final 3D voxels will suffer from the problem of confusing semantic labels~\cite{song2017semantic}.
A real delta noise case is shown in the bottom half part of Figure~\ref{fig:problem}. For the input RGB ``{classroom}'' image, the depth camera mistakenly estimates the depth value of the ``{table}'' as the depth value of ``{furniture}'' (in (b)). Therefore, the visible surface represented by TSDF shifts from the class of ``{table}'' (marked with blue points) to the class of ``{furniture}'' (marked with orange points in (d)). Based on this, the final estimated 3D voxels (in (f)) also mistakenly estimate the part of the ``{table}'' 
as the ``{furniture}''. In comparison, when our SSC model is trained on the visible surface in (e), which is generated by the correct depth value in (c), as the intermediate supervision, semantic labels for both the ``{table}'' and the ``{furniture}'' can be estimated correctly in (g).

In practice, these two types of noise are randomly mixed together to form a more complex noise~\cite{fu2012kinect,zhang2020feature}. To handle these two noise types, although some recent SSC attempts have been made by rendering the noise-free depth value from 3D voxel ground-truth~\cite{silberman2012indoor,firman2016structured}, they are not of practical use as the 3D voxels ground-truth is still needed in inference. However, they indeed validate the potential that more accurate recognition performance can be achieved using the noise-free depth value~\cite{chen20203d,zhang2019cascaded-ccpnet,wang2022ffnet}. To the best of our knowledge, no prior work focuses on mitigating the noisy depth values in SSC without the use of ground-truth depth values in inference. Therefore, the crux is to transfer the clean knowledge learned from ground-truth depth into the noisy-depth pipeline only during training. So, in inference, we can directly use this pipeline without the need for ground-truth.

In this paper, we propose a Cleaner Self (CleanerS) framework to shield the harmful effects of the two depth noises for SSC. 
CleanerS consists of two networks that share the same network architecture (that is what ``self'' means). The only difference between these two networks is that the depth value of the teacher network is rendered from ground-truth, while the depth value of the student network is inherently noisy. Therefore, the depth value of the teacher network is cleaner than the depth value of the student network. 
In the training stage, we make the teacher network to provide intermediate supervision for learning of the student network via knowledge distillation (KD), such that the student network can disentangle the clean visible surface reconstruction and occluded region completion.
To preserve both the detailed information and the abstract semantics of the teacher network, we adopt both \textbf{feature-based} and \textbf{logit-based} KD strategies. In inference, only the student network is used.
Compared to the noisy self, as shown in Figure~\ref{fig:problem}, CleanerS achieves more accurate performance with the help of ground-truth depth values in training but not in testing.
%

\begin{figure*}[!t]
\centering
\includegraphics[width=1.0\textwidth]{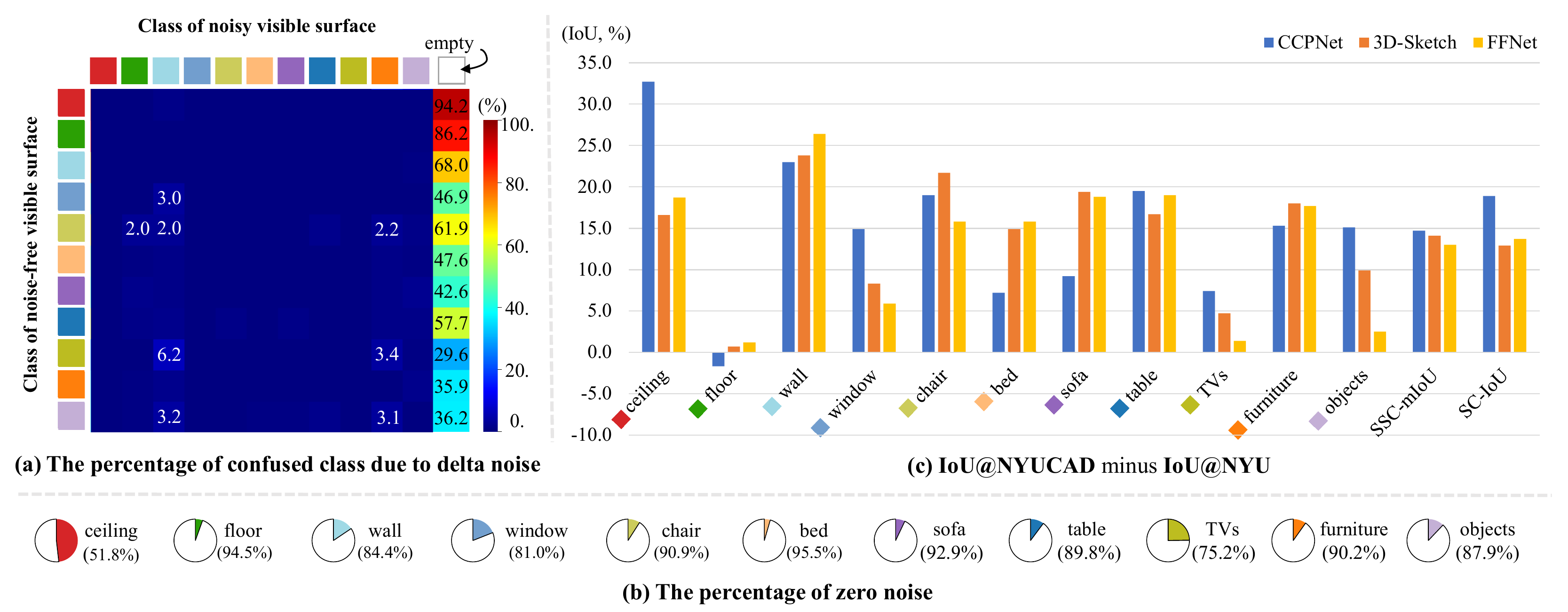}
\vspace{-4mm}
\caption{Quantity of depth noises in NYU~\cite{silberman2012indoor}. (a) the percentage of the delta noise in each object class of the reconstructed visual surface. (b) the percentage of the zero noise on each object class of the depth value. (c) the models' estimation performance gap on the noisy NYU and the noise-free NYUCAD~\cite{firman2016NYUCAD} datasets, where CCPNet~\cite{zhang2019cascaded-ccpnet}, 3D-Sketch~\cite{chen20203d} and FFNet~\cite{wang2022ffnet} are three baselines. 
The performance gap is obtained from results on NYUCAD minus results on NYU.
``IoU'' denotes the evaluation metric of Intersection over Union. 
} 
\label{fig:noise}
\end{figure*}


The main contributions of this work are summarized as the following two aspects: 1) we propose a novel CleanerS framework for SSC, which can mitigate the negative effects of the noisy depth value in training; 2) CleanerS achieves the new state-of-the-art results on the challenging NYU dataset with the input of noisy depth values.

\section{Related Work}
\noindent
\textbf{Semantic Scene Completion.}
SSCNet~\cite{song2017semantic} is the pioneering work to predict both volumetric occupancies and semantic labels for completing 3D scenes.
Follow-up works can be divided into four categories:
1)~\emph{Volume-based methods}. With TSDF as input~\cite{song2017semantic,dourado2019edgenet}, these methods use 3D CNNs for volumetric and semantic predictions.
The disadvantage is the high computational cost of 3D CNNs~\cite{chen20193d,dourado2020semantic}.
2)~\emph{View-volume-based methods}. With RGB image and depth value as inputs, these methods use 2D CNNs to extract features,
and then reshape the features to 3D features by a 2D-3D projection layer for predicting 3D voxels~\cite{guo2018_VVNet,liu2018see,li2019rgbd,li2020attention,liu20203d}. 
These methods are less effective in extracting 3D geometry information due to the limitations of 2D CNNs.
3)~\emph{Point-based methods}. SPCNet~\cite{zhong2020semantic} is a classic method in this category. It particularly resolves the problem of discretization in voxels and predicts SSC by using points as input.  
However, it is sensitive to point-label misalignment (\ie, delta noise).
4)~\emph{Hybrid methods}. Many recent works~\cite{rist2021semantic,rist2020scssnet,cai2021semantic,tang2022not} take at least two kinds of inputs from the set of RGB image, depth value, TSDF, and points. 
They use dual-branch network architectures for feature fusion and achieve state-of-the-art performance.
Our work belongs to this category using RGB image and TSDF as input.
We highlight that our key difference with the above works is that we are the first to particularly resolve the problem of noisy TSDF estimated from noisy depth values.
%

\noindent
\textbf{Learning with Noises.}
%
Both image collection and image annotation bring \emph{data noise} and \emph{label noise}  to computer vision datasets, respectively~\cite{rolnick2017deep,brummer2019natural}. 
Specifically, \emph{data noises} are usually caused by the defects of sensors (\eg, limited visual fields~\cite{henry2014rgb} and inaccurate cameras~\cite{mallick2014characterizations}) or data collection environments (\eg, fog~\cite{gibson2013fast}, rain~\cite{chen2022unpaired}, and nighttime~\cite{rivera2012content}).
To reduce the negative effects on the learning of models, the mainstream idea is to learn clean~\cite{xie2019feature} and robust~\cite{li2021learning,adeli2018semi} feature representations~\cite{zhang2020causal,zhang2020feature}. 
Existing methods can be divided into two categories: noise-cleaning based methods~\cite{ollion2021joint,zheng2021adaptive,liang2021swinir} and robust-modeling based methods~\cite{yang2022treatment,yang2019causal,momeny2021convolutional}.
In particular, the existing models tend to result in unsatisfactory performance when the noises are unpredictable. 
Our method aims to address the problem of noisy data for SSC. 
We propose to transfer the clean knowledge learned from the ground-truth depth value into the noisy-depth pipeline during training. In the inference phase, we can directly use this pipeline without ground-truth. 
%


\noindent
\textbf{Knowledge Distillation (KD).}
KD is proposed in ~\cite{hinton2015distilling} to transfer the knowledge from a teacher model (a stronger or larger model) to a student model (a weaker or smaller model).
The original motivation is model compression.
After KD, the student can achieve a competitive or even superior performance~\cite{li2020few,zhang2021self}.
KD can be roughly categorized into 1) feature-based~\cite{jung2021fair,ji2021show}, logit-based~\cite{zhao2022decoupled,kim2021distilling}, and hybrid~\cite{gou2021knowledge,hsu2022closer}. In this work, we leverage KD for a new purpose: denoising in the SSC task. The teacher model learns cleaner knowledge than the student model with noisy input. This enables the teacher model to provide intermediate supervision for learning the student model via KD.
%
\section{Depth Noises in SSC} 
\label{sec:noise}
In this section, we first compute the percentage of the two noise types: zero noise and delta noise, taking the standard datasets NYU~\cite{silberman2012indoor} and NYUCAD~\cite{firman2016NYUCAD} as examples.

Then, we show the performance gap between the learned models \emph{with} and \emph{without} these noises.
The aim is to quantitatively demonstrate that these two kinds of noises ignored by the existing work of SSC are unfortunately making severe negative effects on the learning of SSC models.
%

\noindent
\textbf{The Quantity of Depth Noises.}
In Figure~\ref{fig:noise} (b) and Figure~\ref{fig:noise} (a), we show the class-wise quantities of zero noises and delta noises, respectively, on NYU~\cite{silberman2012indoor} (including both training and test sets). Each class is represented by a unique color, and the correspondence between colors and class labels is given in Figure~\ref{fig:noise} (c).
%
The way of calculating the percentage of the zero noise on each semantic class of the depth value is expressed as:
\begin{equation} 
\label{eq:zeronoise}
Zero(c) = \frac{\sum \mathbbm{1}_{(Y(d)=c, d \neq 0, d'=0)}}{\sum \mathbbm{1}_{(Y(d)=c, d \neq 0)}},
\end{equation}
where $\mathbbm{1}$ is an indicator (\ie, when conditions in the right parentheses are all satisfied, the current value is incremented by one). $d'$ and $d$ denote the noisy and the noise-free depth value, respectively. $Y(d)$ is the class label of the corresponding visible surface whose 3D position is decided by $d$. $c$ denotes a certain semantic class. The zero noise rate of each class is shown as a pie chart in Figure~\ref{fig:noise} (b).
The way of calculating the percentage of delta noise between $c$ (class of the clean visible surface) and $c'$ (class of the noisy visible surface) is formulated as:
\begin{equation} 
\label{eq:deltanoise}
Delta(c,c') = \frac{\sum \mathbbm{1}_{(Y(d)=c, (d \cdot d') \neq 0, Y(d') = c', c \neq c')}}{\sum \mathbbm{1}_{(Y(d)=c, (d \cdot d') \neq 0)}}.
\end{equation}
%
It is visualized as a block ($c$ row and $c'$ column) on the confusion matrix in Figure~\ref{fig:noise} (a). 

We can observe from Figure~\ref{fig:noise} (b) that all classes in NYU~\cite{silberman2012indoor} have surprisingly significant zero noise rates, \ie, between $51.8\%$ and $95.5\%$.
This means the model trained on such data might be biased by the incomplete visible surface and resulted in an incomplete prediction.
Besides, from Figure~\ref{fig:noise} (a), we can observe that the main semantic confusion (caused by delta noises) is between any semantic class and the special class ``{empty}'', ranging from $29.6\%$ (between ``{empty}'' and ``{TVs}'') to $94.2\%$ (between ``{empty}'' and ``{ceiling}"). Besides, among the non-empty semantic classes of the noisy visible surface, we find that: 1)~several classes confused with the ``{wall}" class; 2)~the class ``{chair}" is easily confused with other non-empty classes. 
The reasons for these phenomena may be that: 1) NYU is an indoor dataset, the visible surfaces of objects are easily shifted to the wall; 2) the chair is not face-structured, and it is difficult to obtain accurate depth values.

\CheckRmv{
\begin{figure*}[!t]
\centering
\includegraphics[width=0.99\textwidth]{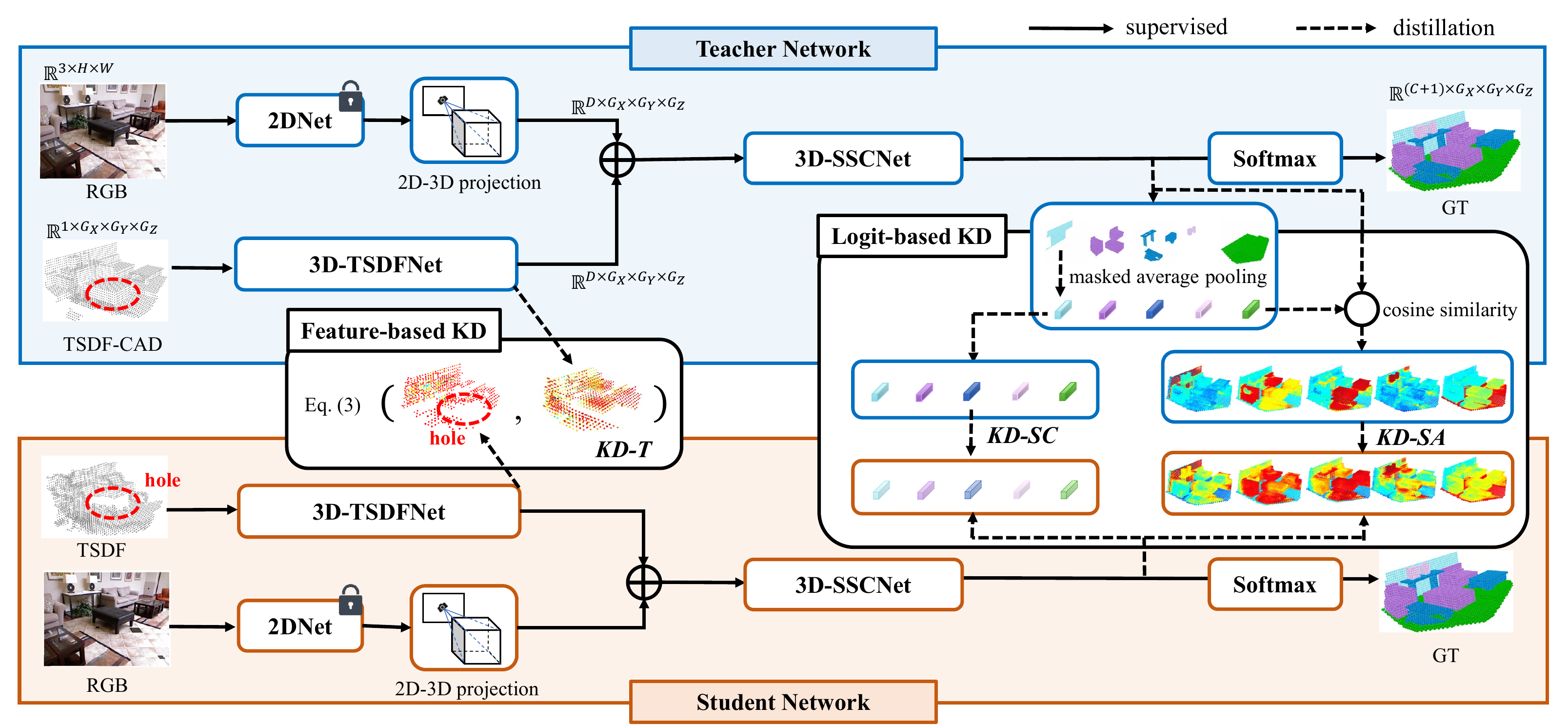}
\caption{Overall architecture of our proposed CleanerS, consisting of two networks: a teacher network, and a student network. The two networks share the same architectures but have different weights. The distillation pipelines include a feature-based cleaner surface distillation (\ie, \emph{KD-T}), and logit-based cleaner semantic distillations (\ie, \emph{KD-SC} and \emph{KD-SA}). The dimensions of the inputs and outputs in the student network are omitted as they are the same as in the teacher network.}
\label{fig:framework}
\end{figure*}
}

\noindent
\textbf{The Effects of Depth Noises.} 
Figure~\ref{fig:noise} (c) shows the performance gap, which is is obtained from the experimental results on the noise-free NYUCAD dataset minus the experimental results on the noisy NYU dataset.
We choose three top-performing methods, CCPNet~\cite{zhang2019cascaded-ccpnet}, 3D-Sketch~\cite{chen20203d} and FFNet~\cite{wang2022ffnet} as the baseline models. 
Each bar denotes a specific method, \eg, the blue bar for CCPNet~\cite{zhang2019cascaded-ccpnet}, the orange bar for 3D-Sketch~\cite{chen20203d}, and the yellow bar denotes FFNet~\cite{wang2022ffnet}, separately. 
For each method, we present the difference of IoU according to the results reported in its paper, which are respectively trained on the NYUCAD~\cite{firman2016NYUCAD} and NYU~\cite{silberman2012indoor} datasets.
In addition to IoU for each semantic class, SSC mean IoU (SSC-mIoU) and SC-IoU are also used as overall evaluation metrics.
%
As in Figure~\ref{fig:noise} (c), most of the performance gaps on IoU exceed $10\%$, and some of them even exceed $20\%$, such as CCPNet~\cite{zhang2019cascaded-ccpnet} in ``{ceiling}'', and FFNet~\cite{wang2022ffnet} in ``{wall}'', which are significant for SSC.
We can also observe that there is a large performance gap between the experimental results \emph{with} or \emph{without} noise from SSC-mIoU and SC-IoU. The above quantitative analysis validates that the depth noises have indeed caused substantial damage to the experimental results of SSC models.
%
%

\section{Cleaner Self (CleanerS)}
In this section, we introduce the implementation details of our approach CleanerS.
Section~\ref{ssec:framework} presents the overall framework of CleanerS. 
Section~\ref{ssec:distillation} elaborates on the feature-based as well as the logit-based KD pipelines implemented in CleanerS. 
Section~\ref{ssec:loss} introduces the overall training losses used in CleanerS.

\subsection{Overall Framework} 
\label{ssec:framework}
Figure~\ref{fig:framework} illustrates the overall framework of CleanerS, consisting of two networks: a teacher network $T$, and a student network $S$. The two networks share the same architectures of 2D and 3D processing modules.
The training process of CleanerS includes the following two steps. \emph{Step 1} is a fully supervised training of $T$ on noise-free depth values that are rendered from 3D voxels ground-truth. \emph{Step 2} includes the fully supervised training of $S$ on noisy depth values, as well as feature-based and logit-based KD training between teacher $T$ and student $S$. 

Specifically, for each network, the {input} consists of an RGB image and a TSDF volume (noise-free TSDF-CAD for $T$, and noisy TSDF for $S$). The outputs are 3D voxels, each of which contains volumetric occupancy and semantic labels. 
In the following, we introduce the three modules in the network: 
1)~Image feature extraction and reformation module. This module has a 2D network to extract image features (a D-channel 2D feature $\mathbf{F}_r$) and a 2D to 3D projection layer that translates 2D feature to 3D feature $\mathbf{V}_r \in \mathbb{R}^{D \times G_X \times G_Y \times G_Z}$. $G_X,G_Y,G_Z$ are the sizes of 3D voxels. The 2D to 3D projection layer is used to lift the feature vector at 2D position $(u,v)$ in $\mathbf{F}_r$ to the corresponding 3D position $(x,y,z)$ according to the depth values of all visible surfaces (otherwise filled by zero vectors).
2)~TSDF feature extraction module. This module transforms the 1-channel 3D input TSDF into a D-channel 3D feature $\mathbf{V}_t$, where $\mathbf{V}_t \in \mathbb{R}^{D \times G_X \times G_Y \times G_Z}$.
3)~SSC prediction module. This module is used to fuse both features $\mathbf{V}_r$ and $\mathbf{V}_t$ by element-wise addition, and then makes predictions via 3D-SSCNet~\cite{chen20203d}. 
The prediction is denoted as $\hat{\mathbf{Y}} \in \mathbb{R}^{(C + 1) \times G_X \times G_Y \times G_Z}$, where $C$ is the number of classes in the dataset, ``$+1$" is for the volumetric occupancy, \ie ``empty" or not.
%

\subsection{Distilling ``Cleaner'' Knowledge} 
\label{ssec:distillation}
%
In the following, we introduce the implementation details of {feature-based} KD and {logit-based} KD.

\noindent
\textbf{Feature-Based KD.} 
%
We use TSDF features for feature-based KD, as follows,
%
\begin{equation} 
\label{eq:distsdf}
L_{KD-T} = MSE(\mathbf{V}_t^S, \mathbf{V}_t^T),
\end{equation}
where $MSE$ denotes mean square error. $\mathbf{V}_t^S$ and $\mathbf{V}_t^T$ are the TSDF features output by the 3D-TSDFNets of student and teacher, respectively.
For visualization (Figure~\ref{fig:framework}), these features are flattened and normalized with a softmax layer along the spatial dimension, and then re-scaled their values into $(0,1)$. In the ``Feature-based KD'' block of Figure~\ref{fig:framework}, we visualize the features of large values ($>0.1$). 
Intuitively, minimizing $L_{KD-T}$ encourages the student to learn to complete the visible ``hole'' in $\mathbf{V}_t^S$ (as shown by the red dashed circle in Figure~\ref{fig:framework}). 

\noindent
\textbf{Logit-Based KD.}
Teacher and student have different inputs, \ie, clean TSDF-CAD \vs noisy TSDF. This results in a large output gap. It is thus ineffective if 
forcing straightforward distillation between teacher's and student's prediction logits. We show an empirical validation in Section~\ref{ssec:ablation}.
On the other hand, teacher and student have the common RGB image input, and aim to predict 3D voxels of the same semantic scene.

Motivated by these, we design two strategies, Semantic-Centered distillation (\emph{KD-SC}) and Semantic Affinity distillation (\emph{KD-SA}), to distill clean semantic knowledge from teacher to student via their prediction logits. 
\emph{KD-SC} is a semantic-centered distillation to transfer global semantic knowledge (from teacher to student).
But it ignores local structures.
\emph{KD-SA} is complementary to \emph{KD-SC}. It is a voxel-to-center affinity distillation to transfer the local structure knowledge. 
Both of them are based on prediction logits, denoted as $\bar{\mathbf{Y}}^T$ (teacher) and $\bar{\mathbf{Y}}^S$ (student), as elaborated in the following.

\noindent
\textbf{KD-SC.}
Given an input sample, we use its ground-truth $\mathbf{Y}$ to generate a binary foreground mask $\mathbf{M}_c \in \mathbb{R}^{G_X \times G_Y \times G_Z}$ for each class $c$. In $M_c$, each voxel stores $1$ if its semantic label is $c$, and otherwise $0$.
Then, we distill the semantic-centered logits for each sample by using the loss:
\begin{equation} 
\label{eq:lossscenter}
L_{KD-SC} = \frac{1}{N_c} \sum_{c=0}^{C} KL(\bar{\mathbf{y}}_c^S, \bar{\mathbf{y}}_c^T),
\end{equation}
where $KL$ is the Kullback–Leibler divergence. $N_c \leq (C + 1)$ is the number of classes appearing on the input sample. $\bar{\mathbf{y}}_c^S$ and $\bar{\mathbf{y}}_c^T$ are the semantic-centered logits of class $c$ in $S$ and $T$, respectively
These logits are calculated based on the prediction logits by applying a masked average pooling along the spatial dimension. E.g., in the student network, they can be derived as:
%
\begin{equation} 
\label{eq:scenter}
\bar{\mathbf{y}}^S_c = \frac{\sum_{i=1}^ {G_X \cdot G_Y \cdot G_Z} (\bar{\mathbf{Y}}^S[i] \cdot \mathbf{M}_c[i])}{\sum_{i=1}^ {G_X \cdot G_Y \cdot G_Z} \mathbf{M}_c},
\end{equation}
where $i$ is the voxel index along the spatial dimension. Due to delta noises, $\bar{\mathbf{y}}_c^S$ will include certain features of the voxels that do not belong to the class $c$ into the average pooling. This results in noisy semantic-centered logits.
Similarly, we can calculate $\bar{\mathbf{y}}_c^T$ for the teacher network, which is ``cleaner'' than $\bar{\mathbf{y}}_c^S$.

\noindent
\textbf{KD-SA.}
To encode local structures, we calculate the voxel-to-center affinity matrices between every voxel and the semantic-centered logits, and then conduct the distillation from $T$ to $S$, as follows,
%
\begin{equation} 
\label{eq:lossaffinity}
L_{KD-SA} = \frac{1}{N_c} \sum_{c=0}^C MSE(\mathbf{A}_c^S, \mathbf{A}_c^T),
\end{equation}
where $\mathbf{A}_c^S$, and $\mathbf{A}_c^T$ are the voxel-to-center affinity matrices, which are calculated in the student and teacher networks, respectively.
E.g., in the student network, the affinity score $a_c^S \in \mathbf{A}_c^S$ is calculated between the semantic-centered logits $\bar{\mathbf{y}}_c^S$ and the logits of each voxel $\bar{\mathbf{y}}^S \in \bar{\mathbf{Y}}^S$ as:
\begin{equation} 
\label{eq:saffinity}
a_c^S = cos(\bar{\mathbf{y}}_c^S, \bar{\mathbf{y}}^S),
\end{equation}
where $cos$ denotes cosine similarity, $\mathbf{A}_c^S \in \mathbb{R}^{G_X \times G_Y \times G_Z}$ is the resultant affinity matrix. The higher affinity value indicates a stronger semantic correlation between the voxel and the center of its label class, and vice versa. 
%

\begin{table*}[t]
\footnotesize
\renewcommand\arraystretch{1.3} 
\setlength{\tabcolsep}{4.0pt}{ 
\begin{center}
\begin{tabular}{r|c|c c c|c c c c c c c c c c c c} 
\hline
\multirow{2}{*}{Methods} & \multirow{2}{*}{Inputs} & \multicolumn{3}{c|}{Scene Completion IoU(\%)} & \multicolumn{12}{c}{Semantic Scene Completion mIoU(\%)} \\ \cline{3-17}
 &  & Prec. & Recall & IoU & \cellcolor{rgb1}ceil. & \cellcolor{rgb2}floor & \cellcolor{rgb3}wall & \cellcolor{rgb4}win. & \cellcolor{rgb5}chair & \cellcolor{rgb6}bed & \cellcolor{rgb7}sofa & \cellcolor{rgb8}table & \cellcolor{rgb9}TVs & \cellcolor{rgb10}furn. & \cellcolor{rgb11}objs. & avg. \\ 
\hline
SSCNet~\cite{song2017semantic} & TSDF & 57.0 & \textbf{94.5} & 55.1 & 15.1 & 94.7 & 24.4 &  0.0 & 12.6 & 32.1 & 35.0 & 13.0 &  7.8 & 27.1 & 10.1 & 24.7\\
ForkNet~\cite{wang2019forknet} & TSDF & - & - & 63.4 & 36.2 & 93.8 & 29.2 & 18.9 & 17.7 & 61.6 & 52.9 & 23.3 & 19.5 & 45.4 & 20.0 & 37.1 \\
CCPNet~\cite{zhang2019cascaded-ccpnet} & TSDF & 74.2  & 90.8 & 63.5 & 23.5 & \textbf{96.3} & 35.7 & 20.2 & 25.8 & 61.4 & \textbf{56.1} & 18.1 & 28.1 & 37.8 & 20.1 & 38.5\\ 
\hline
DDRNet~\cite{li2019rgbd} & RGB+D & 71.5  & 80.8 & 61.0 & 21.1 & 92.2 & 33.5 & 6.8 & 14.8 & 48.3 & 42.3 & 13.2 & 13.9 & 35.3 & 13.2 & 30.4\\ 
AIC-Net~\cite{li2020anisotropic}  & RGB+D &62.4&91.8& 59.2& 23.2 & 90.8 & 32.3 & 14.8 & 18.2  & 51.1  & 44.8 & 15.2   & 22.4 & 38.3  & 15.7  & 33.3 \\
IPF-SPCNet~\cite{zhong2020semantic} & RGB+Point &  70.5  & 46.7  & 39.0 & 32.7  & 66.0 & 41.2 & 17.2 & 34.7 & 55.3  & 47.0 & 21.7 & 12.5  & 38.4  & 19.2  & 35.1 \\
IMENet~\cite{li2021imenet} & RGB+D & \textbf{90.0} & 78.4 & 72.1 & 43.6 & 93.6 & 42.9 & 31.3 & 36.6 & 57.6 & 48.4 & 32.1 & 16.0 & \textbf{47.8} & \textbf{36.7} & 44.2 \\ 
FFNet~\cite{wang2022ffnet}$^{\dagger}$ & RGB+D+TSDF & 89.3 & 78.5 & 71.8 & 44.0 & 93.7 & 41.5 & 29.3 & 36.2 & 59.0 & 51.1 & 28.9 & 26.5 & 45.0 & 32.6 & 44.4 \\
\hline
3D-Sketch~\cite{chen20203d}$^{\dagger}$ & RGB+TSDF & 85.0 & 81.6 & 71.3 & 43.1 & 93.6 & 40.5 & 24.3 & 30.0 & 57.1 & 49.3 & 29.2 & 14.3 & 42.5 & 28.6 & 41.1 \\ 
CleanerS(Res50$^{\dagger}$)  & RGB+TSDF & 89.9	& 79.6	& 73.1	& \textbf{48.2}	& 94.0	& 41.5	& 29.9	& 34.4	& 60.7	& 49.7	& 30.9	& 33.9	& 42.7	& 31.0	& 45.2 \\
CleanerS$^{\star}$  & RGB+TSDF & 88.0	& 83.5	& \textbf{75.0}	& 46.3	& 93.9	& \textbf{43.2}	& \textbf{33.7}	& \textbf{38.5}	& \textbf{62.2}	& 54.8	& \textbf{33.7}	& \textbf{39.2}	& 45.7	& 33.8	& {\bfseries 47.7} \\
\hline
\end{tabular}
\vspace{-2mm}
\caption{Result comparisons with the state-of-the-art methods on the test set of NYU~\cite{silberman2012indoor} (\emph{take the noisy depth value as input}). Results with ``${\dagger}$" and ``${\star}$" denote that these results are based on ResNet50~\cite{he2016deep} and Segformer-B2~\cite{xie2021segformer} as the backbone network for RGB image feature extraction, respectively. Bold numbers represent the best performance.} 
\vspace{-5mm}
\label{tab:sota}
\end{center}}
\end{table*}

\subsection{Overall Loss} 
\label{ssec:loss}
The overall loss of CleanerS consists of two parts: SSC loss and KD loss. 
The SSC loss is applied in both teacher and student networks. First, we minimize the difference between prediction $\hat{\mathbf{Y}}$ and ground-truth $\mathbf{Y}$. Then, we incorporate the following additional supervision for 2DNet. For the 2D feature $\mathbf{F}_r$, we add a $3\times3$ convolution for 2D semantic prediction, which outputs $\hat{\mathbf{Y}}_{2D} \in \mathbb{R}^{(C + 1) \times H \times W}$. 
The 2D semantic ground-truth is noted as $\mathbf{Y}_{2D}$, the class of each pixel at 2D coordinate $[u,v]$ is $\mathbf{Y}[x,y,z]$, where $[x,y,z]$ is the corresponding 3D position after 2D-3D projection. 
Hence, the SSC loss is formulated as:
\begin{equation} 
\label{eq:lossSSC}
L_{SSC} = SCE(\hat{\mathbf{Y}}, \mathbf{Y}) + \lambda SCE(\hat{\mathbf{Y}}_{2D},\mathbf{Y}_{2D}),
\end{equation}
where $SCE$ denotes smooth cross entropy loss~\cite{szegedy2016rethinking}, and $\lambda$ is a coefficient to balance between 3D and 2D semantic losses.
The KD loss is applied in the student network. 
Therefore, the overall loss for the student network can be formulated as:
\begin{equation} \label{eq:lossall}
L_{all}^S = L_{SSC} + \beta \cdot (L_{KD-T} + L_{KD-S}),
\end{equation}
where $L_{KD-S} = L_{KD-SC} + L_{KD-SA}$ is the logit-based KD loss, and $\beta$ is a coefficient to balance between SSC and distillation losses.

\section{Experiments}
\subsection{Datasets and Evaluation Metrics} \label{ssec:datametric}
\noindent\textbf{Datasets.} 
Following~\cite{song2017semantic,Garbade2018_twoStream,chen20203d}, we use the challenging NYU~\cite{silberman2012indoor} dataset and its variants NYUCAD~\cite{firman2016NYUCAD} dataset in our experiments. 
NYU~\cite{silberman2012indoor} consists of 1,449 RGB-D images (795 for training, and 654 for testing) captured via a Kinect sensor~\cite{zhang2012microsoft}. NYU is composed of mainly office and house room scenes. The 3D voxel ground-truth is generated by voxelizing the computer-aided design (CAD) mesh annotations provided by \cite{guo2013support}.
The difference between the two datasets is that the depth values of NYUCAD are rendered from 3D voxel ground-truth, but the depth values of NYU are noisy.
We use the training set of NYUCAD to train the teacher network, and use training and testing sets of NYU to respectively train and evaluate the student network.

\noindent\textbf{Evaluation Metrics.}
Following~\cite{song2017semantic,Garbade2018_twoStream,chen20203d}, we use Precision (Prec.), Recall, and Intersection over Union (IoU) as evaluation metrics for two tasks: semantic scene completion (SSC), and scene completion (SC). 
For SSC, we evaluate the IoU of each category on both observed and occluded voxels, and mean IoU across all categories. 
For SC, we treat every voxel as a binary prediction task, \ie, empty or non-empty, and we evaluate the performance on the occluded 3D voxels.

\subsection{Implementation Details}
\noindent\textbf{Network Architectures.} 
For 2DNet, we adopt the ``B2'' Segformer~\cite{xie2021segformer} as baseline, and initialize it with ImageNet~\cite{deng2009imagenet} pretrained weights. 
For 3D-TSDFNet, we follow 3D-Sketch~\cite{chen20203d} to use the following architecture: three layers of 3D convolution; two DDR blocks~\cite{li2019rgbd} with feature downsampling by a rate of $4$; and two layers of 3D deconvolution for upsampling.
For 3D-SSCNet, we use the same architecture as 3D-TSDFNet except that we remove the first three layers of 3D convolution. 
\emph{Please refer to the supplementary materials for the detailed architecture of each net.}
The feature channel size in both 3D-TSDFNet and 3D-SSCNet is set to $D=256$, and the 3D resolution of prediction $(G_X,G_Y,G_Z)$ is set to $(60,36,60)$. 

\noindent\textbf{Training Settings.} We implement all experiments in this work on the PyTorch~\cite{paszke2019pytorch} with 2 NVIDIA 3090 GPUs. Our model is trained by AdamW~\cite{loshchilov2017decoupled} with a weight decay of 0.05. The batch size is $4$ and the learning rate is initially set as $0.001$ and scheduled by a cosine learning rate decay policy~\cite{loshchilov2016sgdr} with a minimum learning rate of $1e$-$7$. We train our model for $100$ epochs. For each input RGB image, we apply resize, random cropping, and random flipping as 2D data augmentation. For 3D data augmentation, we use random x-axis and z-axis flipping as in \cite{dourado2022data}. In our loss function, both $\lambda$ and $\beta$ are set as $0.25$.

\subsection{Comparisons with State-of-the-Arts}
%
Table~\ref{tab:sota} shows the quantitative results of our CleanerS compared to state-of-the-art methods on the test set of NYU. 
We can see that CleanerS outperforms these methods by clear margins. 
In particular, it achieves the improvements of $3.2\%$ SC-IoU and $3.3\%$ SSC-mIoU compared to the top-performing method FFNet~\cite{wang2022ffnet}, and becomes the new state-of-the-art.
Another closely related work is 3D-Sketch~\cite{chen20203d}. It is also a voxel-based framework with the inputs of RGB image and TSDF.
However, 3D-Sketch uses ResNet50 for 2D feature extraction.
To compare with it fairly, we implement the 2DNet in our method with ResNet50~\cite{he2016deep}, and name it as CleanerS (Res50) in Table~\ref{tab:sota}. 
Results show that our CleanerS (Res50) outperforms 3D-Sketch by $1.8\%$ SC-IoU and $4.1\%$ SSC-mIoU.
%
%

\CheckRmv{
\begin{figure*}[!t]
\centering
\includegraphics[width=0.99\textwidth]{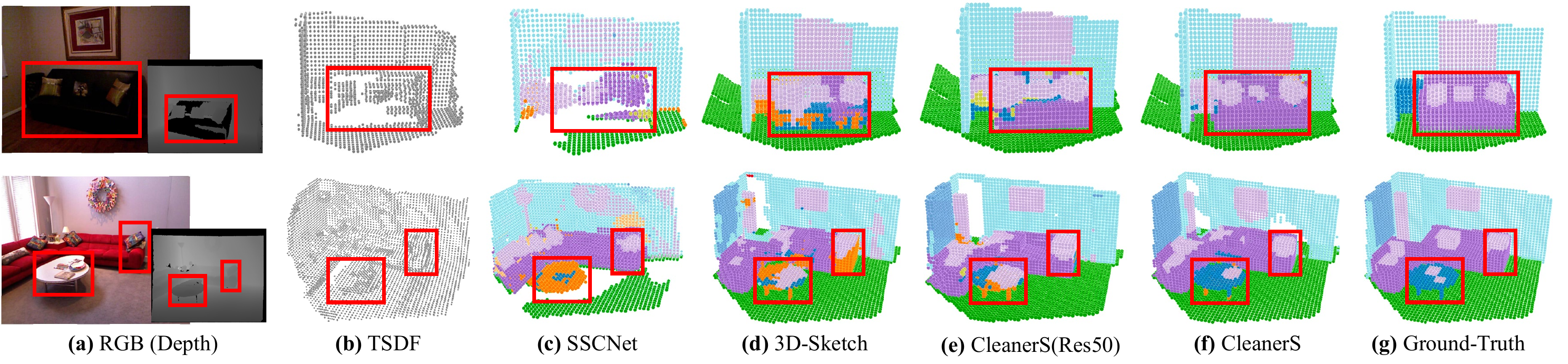}
\caption{Qualitative comparisons with state-of-the-art methods, including SSCNet~\cite{song2017semantic} and 3D-Sketch~\cite{chen20203d}. We show that 1) the incomplete visible surface will lead to incomplete prediction (in the $1_{st}$ row of (c)); 2) the models trained with the supervision of confused class labels will confuse the object semantics in inference ($1_{st}$ row (d) and $2_{nd}$ row (c, d)).}
\label{fig:sota}
\end{figure*}
}

\noindent\textbf{Qualitative Results.} 
Figure~\ref{fig:sota} demonstrates the qualitative results compared to the state-of-the-arts methods, including SSCNet~\cite{song2017semantic} and 3D-Sketch~\cite{chen20203d}. 
On the first row, zero noises in depth values cause incomplete surfaces (in (b)).
As a result, both SSCNet and 3D-Sketch can hardly ``imagine'' the objects on the incomplete surface, resulting in an incomplete prediction in SSCNet~\cite{song2017semantic} and wrong semantic predictions in 3D-Sketch\cite{chen20203d}. 
In the second row, delta noises cause confusion about semantic classes. Thus, both SSCNet and 3D-Sketch are confused about the semantics between the ``{furniture}" and the ``{table}", as well as between the ``{furniture}" and the ``{sofa}". 
Compared to them, we explicitly handle these two kinds of noises, and achieve more accurate predictions as shown in (e) and (f).

\subsection{Ablation Study} 
\label{ssec:ablation}
\noindent\textbf{CleanerS Components.}  
In Table~\ref{tab:ablation}, the $1{st}$ row shows the performance of the teacher model with inputs of RGB and TSDF-CAD (\emph{without noise}). 
It can be taken as the upper bound of the student model. 
The $2{nd}$ row shows the performance of the baseline taking \emph{noisy depth values} as inputs. It is the lower bound.
It is clear that there is a great gap between the upper and lower bounds.
The $3{rd}$ row lists the performance of our CleanerS with \textbf{feature-based KD} only. It shows a significant improvement at SC-IoU $(+2.7\%)$, validating that CleanerS helps completing the scene and gets higher occupancy.
The $4{th}$ row shows the performance of our CleanerS with \textbf{logit-based KD} only. The model learns from the semantic-centered logits and semantic affinities between voxels and class centers given by the teacher network.
It achieves a clear improvement, i.e., $1.7$ percentage points on SSC-mIoU, for the semantic prediction task SSC.
It also brings a surprising improvement on SC-IoU ($2.1$ percentage points). The reason is that the TSDF-CAD features are fused for final prediction, and this enables an implicitly cleaner surface distillation.
Combining both feature- and logit-based KDs achieves the best results.
1) it boosts the best SSC mIoU (in $4_{th}$ row) by an extra $0.5\%$. The more accurate completion means more true positives on volumetric occupancy. This provides the model a chance to infer correct semantic labels;
2) it achieves a further gain of $0.4\%$ above the best SC-IoU (in $3_{rd}$ row). The accurate semantic prediction in turn enhances the prediction of shapes of occluded regions. 
Overall, our CleanerS gains the improvements of $3.1\%$ SC-IoU and $2.2\%$ SSC-mIoU, compared to the baseline.

\begin{table}[t]
\footnotesize
\renewcommand\arraystretch{1.3} 
\setlength{\tabcolsep}{.7pt}{ 
\begin{center}
\begin{tabular}{r|c| c c c | c c} 
\hline
Methods & {Inputs} & \textbf{$L_{SSC}$} & \textbf{$L_{KD-T}$} & \textbf{$L_{KD-S}$}   & {{SC-IoU}} & {{SSC-mIoU}} \\ \hline
Teacher & RGB+TSDF-CAD & \cmark &  &  & 85.3\% 	& 59.4\%   \\
\hline 
Baseline & RGB+TSDF & \cmark &  &  & 71.9\%  & 45.5\%  \\
CleanerS & RGB+TSDF & \cmark & \cmark &  & 74.6\%  	& 46.0\%   \\
CleanerS & RGB+TSDF & \cmark &  & \cmark  & 74.0\%  	& 47.2\%  \\
CleanerS & RGB+TSDF & \cmark  & \cmark & \cmark  & 75.0\%  	& 47.7\%  \\
\hline
\end{tabular}
\end{center}}
\vspace{-3mm}
\caption{Ablation study results of different components in our CleanerS on the test set of NYU~\cite{silberman2012indoor}.}
\vspace{-3mm}
\label{tab:ablation}
\end{table}


\noindent\textbf{Logit-based KD \vs Soft-target KD.} As mentioned in the second paragraph of Section~\ref{ssec:distillation}, directly forcing straightforward distillation is ineffective.
To validate this, we compare our logit-based distillation (including $L_{KD-SC}$ and $L_{KD-SA}$) with soft target distillation~\cite{hinton2015distilling}, and we denote the loss of the latter one as $L_{KD-Soft}$. %
From the results in Table~\ref{tab:dis_semantic}, we can observe that the $L_{KD-SC}$ plays a major role, and $L_{KD-Soft}$ achieves similar performance as using $L_{KD-SA}$. 
More specifically, we have the following conclusions: 1)~it is sub-optimal to apply only a local voxel-wise logits distillation, and 2)~it is a better choice to distill the global semantic center embedding. The reasons might be that the inputs to teacher and student models are not exactly the same, clean TSDF-CAD \emph{v.s.} noisy TSDF, and it is hard to match their outputs in a voxel-wise manner. Besides, with the same RGB image input, both teacher and student models aim to infer 3D semantic voxels in the same scene, and it is thus reasonable to enforce the two models to predict similar semantic-centered logits for each semantic class.

\begin{table}[t]
\footnotesize
\renewcommand\arraystretch{1.3} 
\setlength{\tabcolsep}{3pt}{ 
\begin{center}
\begin{tabular}{r | c c c | c c} 
\hline
Methods & \textbf{$L_{KD-SC}$} & {$L_{KD-SA}$} & {$L_{KD-Soft}$}   & {{SC-IoU}} & {{SSC-mIoU}} \\ \hline
Baseline &  &  &  & 71.9\% & 45.5\% \\ \hline
CleanerS    &  &   & \cmark  & 73.7\% 	& 46.3\% \\ \hline
CleanerS    &  & \cmark  &  & 73.6\% 	& 46.3\% \\
CleanerS & \cmark &   &  & 74.0 \%	& 46.7\%  \\
CleanerS & \cmark  & \cmark &   & 74.0\% 	& 47.2\%  \\
\hline
\end{tabular}
\end{center}
\vspace{-3mm}
\caption{The results of different semantic distillation strategies in our CleanerS on the test set of NYU~\cite{silberman2012indoor}.}
\vspace{-3mm}
\label{tab:dis_semantic}}
\end{table}

\section{Conclusions}
We presented a novel CleanerS framework for tackling the task of SSC.
We first computed the percentage of zero noise and delta noise in the depth values. 
Then, we showed that these two types of noises have caused severe negative effects on SSC models. In our CleanerS, we trained a teacher network to learn clean knowledge and provided intermediate supervision for the learning of a student network.
Compared to the existing SSC methods, CleanerS can not only achieve more accurate performance but also requires no ground-truth labels in inference.
Extensive experimental results validated that our CleanerS can shield the negative effects of noisy depth values, and achieve the new state-of-the-art on the challenging NYU dataset.

\section*{Acknowledgements} This research was supported by the A*STAR under its AME YIRG Grant (Project No.A20E6c0101), the National Key Research and Development Program of China under Grant 2018AAA0102002, the National Natural Science Foundation of China under Grant 61925204, and AI Singapore AISG2-RP-2021-022.

{\small
\bibliographystyle{ieee_fullname}
\bibliography{cvpr2023/Main}
}
\clearpage

\beginsupp

\noindent {\Large \textbf{Supplementary Materials}}
\\

This supplementary includes an introduction to the background of 2D-3D projection (\ref{subsec_2d3dprojection}), 
implementation details of both 3D-TSDFNet and 3D-SSCNet (\ref{suppsec_3dnet}), 
ablation results of CleanerS with different 2DNets (\ref{suppsec_2dnet}) and with different resolutions (\ref{suppsec_resolution}), comparison results of using different methods of mitigating the depth noise (\ref{suppsec_denoise}), the correlation between the noise accuracy degrade (\ref{suppsec_noiserate}), and more visualization results (\ref{suppsec_visual}).

\section{2D-3D Projection}
\label{subsec_2d3dprojection}
{\color{red}{This supplementary is for the background introduction of the main paper.}} 
2D-3D projection layer is used to recover the visible surface in a 3D scene such as to map every pixel in a 2D image to its corresponding 3D spatial position. Given the depth image $\mathbf{I}_{d}$, each pixel at 2D position $[u,v]$, with a depth value $d=\mathbf{I}_d(u,v)$, is projected to the 3D position $[x,y,z]$. This mapping $\mathbb{M}$ can be expressed as follows:
\begin{equation} \label{eq:proj}
[x,y,z] = \mathbb{M}(u,v,d).
\end{equation}
%
Specifically, this projection includes the following two steps:

\noindent
\emph{Step 1}: Mapping each 2D pixel to an individual 3D point based on the imaging information including an intrinsic camera matrix $\mathbf{K} \in \mathbb{R}^{3\times3}$ and an extrinsic camera matrix $[\mathbf{R}|\mathbf{t}] \in \mathbb{R}^{3\times4}$. The mapping satisfies the following equation:
\begin{equation} \label{eq:2d3d}
[\mathbf{R}|\mathbf{t}][x_p,y_p,z_p,1]^{\top} = \mathbf{K}^{-1}([u, v, 1]^{\top} \cdot d).
\end{equation}
Based on Eq.~\eqref{eq:2d3d}, we can solve the $[x_p,y_p,z_p]$, i.e., the 3D position of the corresponding point.

\noindent
\emph{Step 2}: Discretizing the point position into a grid voxel with a given unit voxel size $g$,
\begin{equation} \label{eq:discret}
[x,y,z] = [\lfloor x_p / g + 0.5 \rfloor, \lfloor y_p / g + 0.5 \rfloor, \lfloor z_p / g + 0.5 \rfloor],
\end{equation}
where $\lfloor \cdot \rfloor$ is the floor rounding. The unit grid size $g$ is $0.08m$ and the resultant 3D voxel size is $(60, 36, 60)$. 
The 2D-3D projection is used for two purposes in this work: 1) getting the class labels of 2D pixels (resulting $Y(d)$ in Section~\textcolor{red}{3} and $\mathbf{Y}_{2D}$ in Section~\textcolor{red}{4.3}); 2) translating a 2D feature into a 3D feature (translating from $\mathbf{F}_r$ to $\mathbf{V}_r$ in Section~\textcolor{red}{4.1}). 

%
\begin{figure*}[!t]
    \centering
    \includegraphics[width=0.73\textwidth]{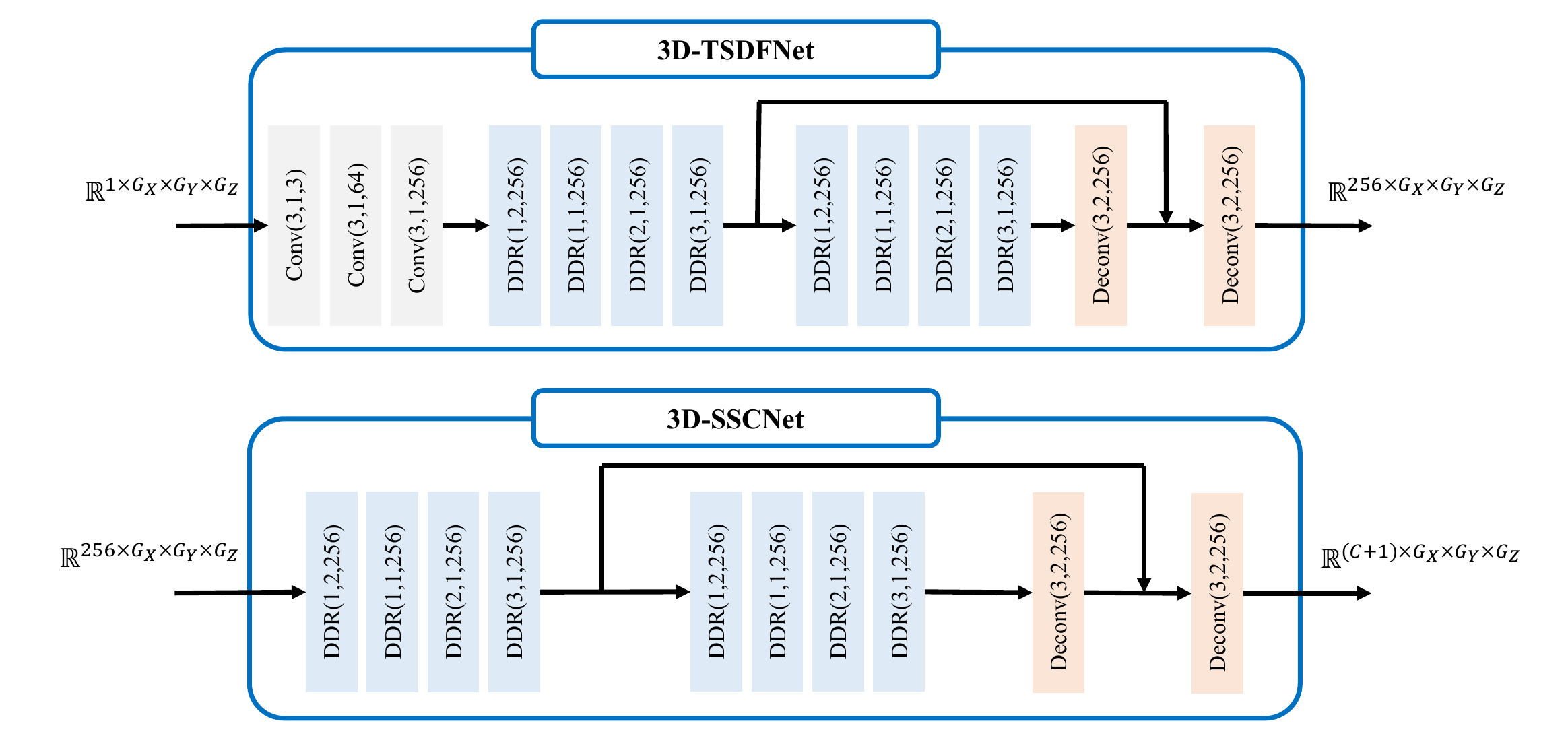}
    \caption{
    The architectures of 3D-TSDFNet and 3D-SSCNet. $\text{Conv}(k,d,c)$ is a 3D convolutional layer with kernel size $k$, dilation $d$, and output channel $c$; $\text{DDR}(d,r,c)$ is a DDR block~\cite{li2019rgbd} with dilation $d$, downsample rate $r$, and output channel $c$; and $\text{Deconv}(k,R,c)$ is a 3D deconvolutional layer with kernel $k$, upsample rate $R$, and output channel $c$.
  } 
\label{fig:3dnet}
\end{figure*}

\section{Implementation Details of 3D-TSDFNet and 3D-SSCNet}
\label{suppsec_3dnet}
{\color{red}{This supplementary is for Section 4.1 of the main paper.}}  We introduce the details of the 3D-TSDFNet and 3D-SSCNet in both student and teacher networks, as presented in Figure~\ref{fig:3dnet}. Following 3D-Sketch~\cite{chen20203d}, the 3D-TSDFNet includes 3 layers of 3D convolutions to encode the input TSDF into high dimensional features, 8 DDR blocks with different dilations and a downsample rate of 4 to enlarge receptive fields and reduce computation costs, and 2 layers of 3D deconvolutions to upsample features to have the same volume size as the input TSDF volume. Besides, a skip connection is added between each pair of DDR block and deconvolution layer for efficient gradient back-propagation. The 3D-SSCNet uses the same architecture as the 3D-TSDFNet except that it removes the first 3 layers of 3D convolutions.

\begin{table}[t]
\footnotesize
\renewcommand\arraystretch{1.3} 
\newcommand{\tabincell}[2]{\begin{tabular}{@{}#1@{}}#2\end{tabular}}
\setlength{\tabcolsep}{.7pt}{ 
\begin{center}
\begin{tabular}{r|c| c | c | c } 
\hline
Methods & {Inputs} & 2DNet  & {{SC-IoU}} & {{SSC-mIoU}} \\ \hline
Baseline & RGB+TSDF & \multirow{3}{*}{ResNet50}  & 70.6\%  & 43.2\%  \\
Teacher & RGB+TSDF-CAD  & & 85.1\% ($\uparrow$ 14.5\%)	& 57.1\%  ($\uparrow$ 13.9\%) \\
CleanerS & RGB+TSDF &  & 73.1\% ($\uparrow$ 2.5\%) 	& 45.2\% ($\uparrow$ 2.0\%) \\
\hline
Baseline & RGB+TSDF & \multirow{3}{*}{\tabincell{c}{Segformer\\-B2}}  & 71.9\%  & 45.5\%  \\
Teacher & RGB+TSDF-CAD  & & 85.3\% ($\uparrow$ 13.4\%)	& 59.4\%  ($\uparrow$ 13.9\%) \\
CleanerS & RGB+TSDF &  & 75.0\% ($\uparrow$ 3.1\%) 	& 47.7\% ($\uparrow$ 2.2\%) \\
\hline
\end{tabular}
\end{center}}
\vspace{-3mm}
\caption{The ablation study results for using different 2DNets (ResNet50~\cite{he2016deep} \vs. Segformer-B2~\cite{xie2021segformer}) in our CleanerS on the test set of NYU~\cite{silberman2012indoor}.}
%
\label{tab:2dnet}
\end{table}

\section{Ablation Results with Different 2DNet}
\label{suppsec_2dnet}
{\color{red}{This supplementary is for Section 5.4 of the main paper.}}  
We supplement the ablation study for CleanerS-Res50 and compare the results with those of CleanerS in Table~\ref{tab:2dnet}. From the table, we can observe that: 
1) For all the methods, including baseline, teacher, and CleanerS, better performances are achieved by using Segformer-B2 (than using ResNet50), especially on the metric of SSC mIoU. The reason is that image features extracted from the transformer-based Segformer-B2 encode better global (i.e., longer-range) context information.
2) With different 2DNet architectures (Segformer-B2 or ResNet50), the proposed CleanerS can consistently improve over baseline by large margins, \ie, over 2.5\% for SC-IoU and over 2.0\% for SSC-mIoU. In addition, using better 2DNets (Segformer-B2), our CleanerS achieves even higher improvement (over baseline). This demonstrates the effectiveness and robustness of the proposed CleanerS for resolving the problem of depth noise in SSC.
\section{Ablation Results with Different Resolutions}
\label{suppsec_resolution}

{\color{red}{This supplementary is for Section 5.4 of the main paper}}. We conduct experiments with a lower resolution under the 3D size of ($30$, $18$, $30$). Experimental results are given in Table~\ref{Rtab:resolution}, where ``HR/LR'' denotes high/low resolution. Furthermore, ``HR2LR'' is a variant by distilling from an HR teacher to an LR student, which is meant to verify if an HR teacher will enable better knowledge distillation. We can observe that: 1) compared to HR, our CleanerS with LR results in a higher SC-IoU and a lower SSC-mIoU; 2) HR2LR performs even worse than LR2LR, which suggests the same resolution inputs enable a better knowledge distillation.

\begin{table}[t]
\footnotesize
\renewcommand\arraystretch{.95} 
\setlength{\tabcolsep}{1.8pt}{ 
\begin{center}
\begin{tabular}{r|c  c | c c | c c} 
\hline
\multirow{2}{*}{Methods}    & \multicolumn{2}{c|}{HR}    & \multicolumn{2}{c|}{LR}     & \multicolumn{2}{c}{HR2LR}\\
         & SC-IoU                   & SSC-mIoU                 & SC-IoU                & SSC-mIoU       & SC-IoU                   & SSC-mIoU          \\
\hline
Baseline   & 71.9\%          & 45.5\%       & 80.1\%    & 38.6\%     & -  & -   \\
CleanerS   & \textbf{75.0\%}          & \textbf{47.7\%}       & \textbf{81.8\%}     & \textbf{40.6\%}    & 81.1\%  & 40.5\% \\
\hline
\end{tabular}
\end{center}}
\vspace{-3mm}
\caption{The ablation study results of our CleanerS with different 3D resolution on the test set of NYU~\cite{silberman2012indoor}.}
\label{Rtab:resolution}
\end{table}

%

\section{Feature-based KD \vs. Data-based Denoise}
\label{suppsec_denoise}
{\color{red}{This supplementary is for Section 5.4 of the main paper.}}  In related works, there is a common practice to mitigate noises by using the corresponding clean data as learning targets~\cite{yang2018proximal,liu2021disentangling}. We validate here that it is not working for our cases. Specifically, we mitigate the noise in TSDF by using the noise-free TSDF-CAD input as a learning target.
First, we add an extra prediction layer after 3D-TSDFNet, which with an input TSDF feature $V_t^S$. The prediction layer includes a DDR block~\cite{li2019rgbd} and a 3D convolutional layer, which outputs a 3-channel prediction (2-channel for the sign prediction and 1-channel for distance prediction).
Then, we compare its results with our feature-based KD, in Table~\ref{tab:dis_tsdf}.
As shown in Table~\ref{tab:dis_tsdf}, it achieves a limited performance gain (0.4\% on SC-IoU and 0.4\% on SSC-mIoU).
In contrast, our feature-based KD significantly improves the SC-IoU. We think there are two reasons. 1) The TSDF-CAD features are task-oriented features and have a richer representation than the TSDF-CAD input. 2) Taking the TSDF-CAD input as a learning target needs extra prediction layers, which might distract the optimization. 
%

\begin{table}[t]
\footnotesize
\renewcommand\arraystretch{1.3} 
\setlength{\tabcolsep}{7pt}{ 
\begin{center}
\begin{tabular}{r | c | c c} 
\hline
Methods & Intermediate Supervision  & {{SC-IoU}} & {{SSC-mIoU}} \\ \hline
Baseline    &  -                   & 71.9 \% 	& 45.5\% \\ \hline
CleanerS    &  TSDF-CAD input       & 72.3 \% 	& 45.9 \% \\ \hline
CleanerS    &  TSDF-CAD feature    & {74.6\%} 	& 46.0\% \\ \hline
\end{tabular}
\end{center}
\caption{The results of using different methods to mitigate noises in the TSDF on the test set of NYU~\cite{silberman2012indoor}. We use either of TSDF-CAD inputs or TSDF-CAD features (output by the teacher network) to be an intermediate supervision in 3D-TSDFNet.}
\label{tab:dis_tsdf}}
\end{table}
%

\section{Correlation between the Noise Rate and Accuracy Degrade}
\label{suppsec_noiserate}

{\color{red}{This supplementary is for Section 5.4 of the main paper}}. To figure out the correlation between the noise rate and accuracy degradation, we perform the synthetic noise depth input by randomly adding either one or both of the zero noise and delta noise into the clean depth-CAD. The noise rate is gradually set to $20\%$, $50\%$, and $80\%$.
Experimental results are given in Table~\ref{Rtab:noiserate}. We can observe that 1) the higher the noise rate, the more degradation of the accuracy; 2) mixing both zero noise and delta noise will result in a complex noise and drops the performance drastically, especially for the SSC-mIoU.

\begin{table}[h]
\footnotesize
\renewcommand\arraystretch{1.3} 
\newcommand{\tabincell}[2]{\begin{tabular}{@{}#1@{}}#2\end{tabular}}
\setlength{\tabcolsep}{7pt}{ 
\begin{center}
\begin{tabular}{r | c | c | c } 
\hline
 & \multicolumn{3}{c}{ SC-IoU/SSC-mIoU (\%)} \\  \hline
 Noise Rate &20\%   &50\%   &80\%   \\  \hline
\textbf{Zero Noise}  & \textcolor{red}{$\downarrow$2.2 / $\downarrow$1.9}   & \textcolor{red}{$\downarrow$2.4 / $\downarrow$2.6} & \textcolor{red}{$\downarrow$2.7 / $\downarrow$3.3} \\  \hline
\textbf{Delta Noise}   &\textcolor{red}{$\downarrow$5.1 / $\downarrow$4.2} & \textcolor{red}{$\downarrow$5.5 / $\downarrow$5.6}   & \textcolor{red}{$\downarrow$5.9 / $\downarrow$6.5} \\ \hline
\textbf{Mix Both Noises}  &\textcolor{red}{$\downarrow$8.7 / $\downarrow$5.0} & \textcolor{red}{$\downarrow$12.6 / $\downarrow$5.5}   & \textcolor{red}{$\downarrow$14.0 / $\downarrow$6.8} \\
\hline
\end{tabular}
\end{center}}
\vspace{-3mm}
\caption{Results of synthetic depth with different noise rates.}
\label{Rtab:noiserate}
\end{table}


\begin{figure*}[h]
\centering
\includegraphics[width=0.99\textwidth]{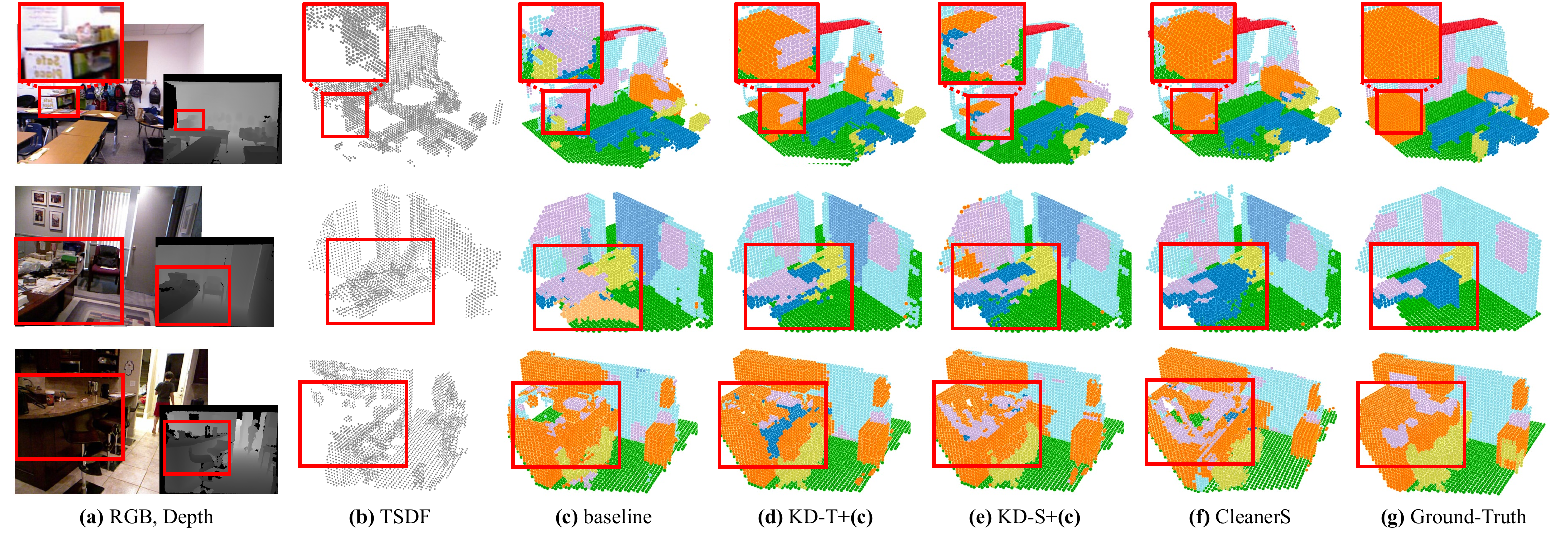}
\caption{Visualization results for ablation study. The proposed feature-based KD (in (d)) and logit-based KD (in (e)) improve the baseline with better volumetric occupancy and semantics. Combining both (in (f)) achieves the best results.} 
\label{fig:ablation}
\end{figure*}

\begin{figure*}[h]
\centering
\includegraphics[width=0.975\textwidth]{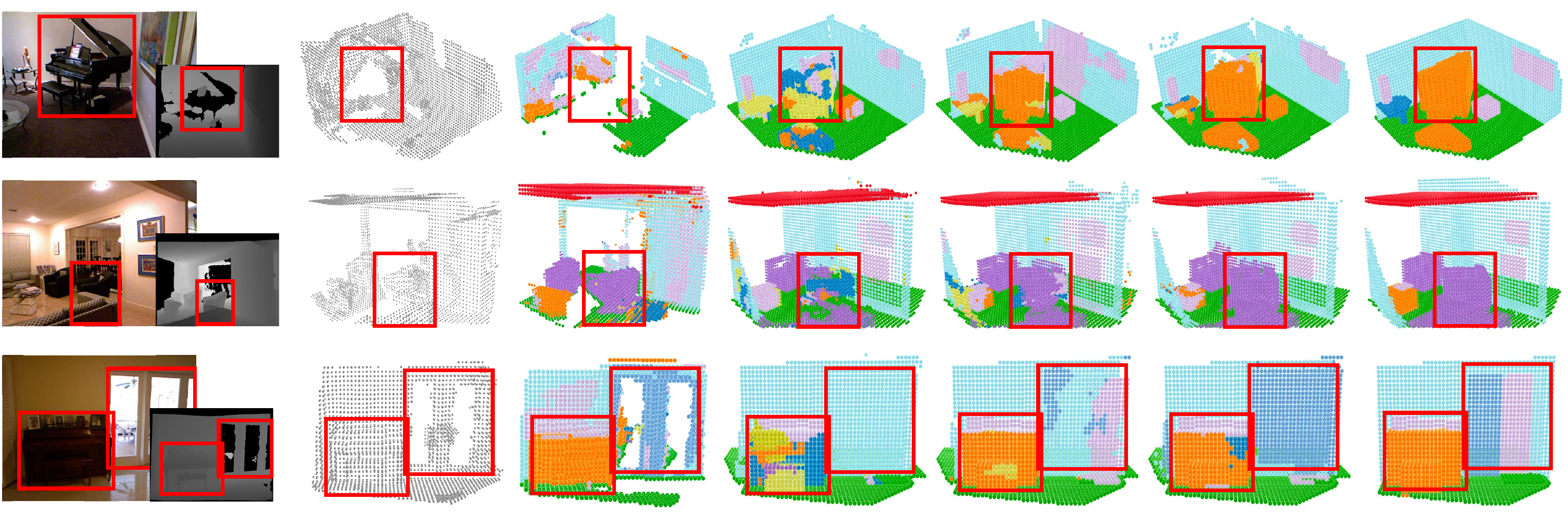}
\includegraphics[width=0.99\textwidth]{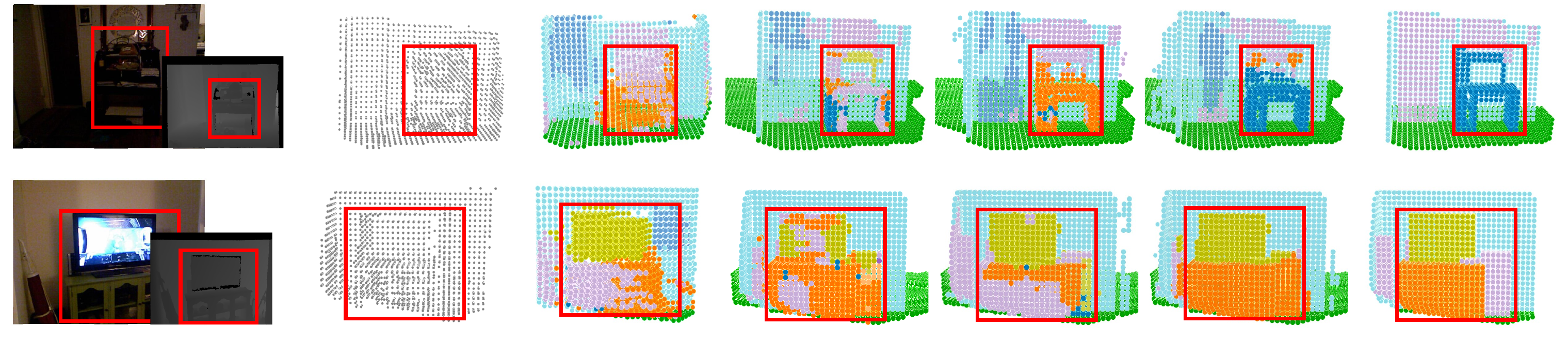}
\caption{More qualitative comparisons with state-of-the-art methods, including SSCNet~\cite{song2017semantic} and 3D-Sketch~\cite{chen20203d}. We present several challenging examples with zero noises and delta noises.}
\label{fig:sota2}
\end{figure*}


\section{More Visualization Results}
\label{suppsec_visual}
{\color{red}{This supplementary is for Section 5.4 of the main paper.}}  As shown in Figure~\ref{fig:ablation}, compared with the baseline method (in (c)), the cleaner surface distillation by feature-based KD (in (d)) helps to get cleaner occupancy predictions but may confuse the semantics. Combining it with the cleaner semantic distillation by logit-based KD (in (f)) can resolve this confusion.
In Figure~\ref{fig:sota2}, we supplement more visual examples compared to state-of-the-art methods.

\end{document}